\documentclass{article}
\usepackage{spconf}
\usepackage{amsmath,amssymb,dsfont,graphicx,verbatim}
\graphicspath{{figures/}}
\usepackage{amsmath,amssymb,dsfont,verbatim}
\usepackage{import}
\usepackage{caption}
\usepackage{subcaption}
\usepackage[ruled,vlined]{algorithm2e}
\usepackage{cite}
\usepackage{xcolor}
\usepackage{hyperref}
\usepackage{booktabs} 
\usepackage{tikz}
\usepackage{siunitx}
\usepackage{balance}
\usetikzlibrary{mindmap}

\allowdisplaybreaks[0]

\usepackage{amsthm}

\DeclareMathOperator*{\argmin}{arg\,min}

\DeclareMathOperator{\T}{\mathsf{T}}
\DeclareMathOperator{\E}{\mathds{E}}

\DeclareMathOperator{\w}{\boldsymbol{w}}
\DeclareMathOperator{\x}{\boldsymbol{x}}
\DeclareMathOperator{\s}{\boldsymbol{s}}

\usepackage{amsthm}

\theoremstyle{plain}

\newtheorem{assumption}{Assumption}
\newtheorem{theorem}{Theorem}

\let\OLDthebibliography\thebibliography
\renewcommand\thebibliography[1]{
  \OLDthebibliography{#1}
  \setlength{\parskip}{5pt}
  \setlength{\itemsep}{0pt plus 0.6ex}
}

\title{Robust and Efficient Aggregation for Distributed Learning}
\name{Stefan Vlaski\(^{\star}\), Christian Schroth\(^{\dagger}\), Michael Muma\(^{\dagger}\) and Abdelhak M. Zoubir\(^{\dagger}\)}
\address{\(^{\star}\)Department of Electrical and Electronic Engineering, Imperial College London, UK \\ \(^{\dagger}\)Signal Processing Group, Technische Universität Darmstadt, Germany\thanks{Emails: s.vlaski@imperial.ac.uk, \{cschroth, mmuma, zoubir\}@spg.tu-darmstadt.de}}

\begin{document}
\maketitle
\begin{abstract}
  Distributed learning paradigms, such as federated and decentralized learning, allow for the coordination of models across a collection of agents, and without the need to exchange raw data. Instead, agents compute model updates locally based on their available data, and subsequently share the update model with a parameter server or their peers. This is followed by an aggregation step, which traditionally takes the form of a (weighted) average. Distributed learning schemes based on averaging are known to be susceptible to outliers. A single malicious agent is able to drive an averaging-based distributed learning algorithm to an arbitrarily poor model. This has motivated the development of robust aggregation schemes, which are based on variations of the median and trimmed mean. While such procedures ensure robustness to outlier\textcolor{black}{s} and malicious behavior, they come at the cost of significantly reduced sample efficiency. This means that current robust aggregation schemes require significantly higher agent participation rates to achieve a given level of performance than their mean-based counterparts in non-contaminated settings. In this work we remedy this drawback by developing statistically efficient and robust aggregation schemes for distributed learning.
\end{abstract}
\section{Introduction and Related Works}\label{sec:intro}
We consider a general distributed learning problem, where a collection of \( K \) agents aim to collaboratively solve a stochastic optimization problem defined through:
\begin{align}\label{eq:main_problem}
  w^o \triangleq \argmin_w \frac{1}{K} \sum_{k=1}^{K} \mathds{E} Q(w; \x_k)
\end{align}
Here, \( \x_k \) denotes a random variable describing the privately available data at agent \( k \), and \( Q(w; \x_k) \) denotes the associated loss. It will be convenient to define \( J_k(w) \triangleq \mathds{E} Q(w; \x_k) \) and \( J(w) \triangleq \sum_{k=1}^{K} p_k J_k(w) \), so that:
\begin{align}
  J(w) = \frac{1}{K}\sum_{k=1}^{K} J_k(w) = \frac{1}{K}\sum_{k=1}^{K} \mathds{E} Q(w; \x_k)
\end{align}
This formulation is general enough to cover a wide range of learning problems, from distributed least mean-squares and logistic regression~\cite{Sayed14} to distributed deep learning~\cite{Lian17, Vlaski19nonconvexP2}.

Solutions to consensus optimization problems of the form~\eqref{eq:main_problem} can be pursued through a number of distributed strategies, depending on resource and communication constraints. Broadly, algorithms for distributed learning can be classified into (a) fusion-center based strategies, and (b) fully-decentralized strategies. Fusion-center based strategies involve communication with a central parameter server, which performs aggregation of intermediate model estimates, and subsequently broadcasts them back to participating agents. Fully-decentralized approaches on the other hand rely purely on peer-to-peer exchanges over some (potentially sparse) graph topology.

\textbf{Example 1 -- Federated learning:} Federated architectures rely on a central processor to coordinate computations, but avoid exchanges of raw data by allowing agents to locally compute updates of a common model in a highly asynchronous manner. A representative example is the federated averaging algorithm~\cite{Konecny16}, where at each iteration \( i \), a subset \( \mathcal{N} \) of \( N \) agents is chosen, and each agent is provided with the current version of the model \( \boldsymbol{w}_{i-1} \) stored at the central parameter server. Each agent then initializes \( \boldsymbol{\phi}_{k, 0} = \boldsymbol{w}_{i-1} \) and performs \( L_k \) steps of (stochastic) gradient descent by iterating over \( j \):
\begin{equation}\label{eq:fed_adapt}
  \boldsymbol{\phi}_{k, j} = \boldsymbol{\phi}_{k, j-1} - \mu \widehat{\nabla J}_k(\boldsymbol{\phi}_{k, j-1})
\end{equation}
\textcolor{black}{Here, \( \mu > 0 \) denotes the step-size and} \( \widehat{\nabla J}_k(\boldsymbol{\phi}_{k, j-1}) \) corresponds to a stochastic gradient approximation of \( J_k(w) \) based on the locally available data. \textcolor{black}{Upon completion, each agent} returns \( \boldsymbol{\phi}_{k, L_k} \) to the parameter server, where the aggregate model is updated according to:
\begin{equation}\label{eq:fed_combine}
  \boldsymbol{w}_i = \frac{1}{N} \sum_{k \in \mathcal{N}} \boldsymbol{\phi}_{k, L_k}
\end{equation}

\textbf{Example 2 -- Decentralized learning:} In contrast to federated approaches, decentralized learning algorithms rely solely on peer-to-peer interactions between pairs of agents connected by some (potentially sparse) graph topology, and avoid the need for a central aggregator or coordinator. Similar to federated structures, these algorithms perform combinations of local updates steps, based on locally available data, and aggregation steps, with the difference being that instead of aggregating at a central processor, aggregation occurs locally over neighborhoods of agents based on peer-to-peer exchanges. Here, the neighborhood \( \mathcal{N}_k \) of agent \( k \) defines the set of agents, with which agent \( k \) is willing and able to exchange information. An example is the ATC-diffusion algorithm, which takes the form~\cite{Sayed14}:
\begin{align}
  \boldsymbol{\phi}_{k,i} &= \w_{k,i-1} - \mu \widehat{\nabla J}_{k}(\w_{k,i-1})\label{eq:adapt}\\
  \w_{k,i} &= \sum_{\ell \in \mathcal{N}_k} a_{\ell k} \boldsymbol{\phi}_{\ell,i}\label{eq:combine}
\end{align}
Examining relations~\eqref{eq:fed_combine} and~\eqref{eq:combine}, we note that both federated and decentralized learning approaches rely on an averaging step of the form:
\begin{align}\label{eq:general_aggregation}
  \w_{k, i} = \sum_{\ell \in \textcolor{black}{\mathcal{N}_k}} a_{\ell k} \boldsymbol{\phi}_{\ell, i} = \argmin_{\w} \sum_{\ell \in \mathcal{N}_k} a_{\ell k} {\|\boldsymbol{\phi}_{\ell, i} - \w\|}^2
\end{align}
for some non-negative weights \( a_{\ell k} \) that add up to one. This immediately makes clear the limited robustness of averaging-based schemes for distributed learning. Manipulating the value of a single \( \boldsymbol{\phi}_{\ell, i} \), either for benign or malicious reasons, has the potential to influence the aggregate model \( \w_{k, i} \) arbitrarily. This has motivated increased interest over recent years on robust alternatives to the aggregation scheme~\eqref{eq:general_aggregation}. An example is the secure aggregation protocol of~\cite{Kakade19} based on the geometric median \textcolor{black}{(also known as spatial median)}, which takes the form:
\begin{align}\label{eq:median_aggregation}
  \w_{k, i} = \argmin_{\w} \sum_{\ell \in \mathcal{N}_k} a_{\ell k} \|\boldsymbol{\phi}_{\ell, i} - \w\|
\end{align}
Variations based on element-wise median/trimmed-mean have also been considered~\cite{Yin18}. The authors of~\cite{Blanchard17} consider a more elaborate procedure termed ``Krum'', which nevertheless discards a majority of (potentially) benign samples. While these approaches yield increased robustness to perturbations in \( \boldsymbol{\phi}_{\ell, i}(m) \) up to a contamination rate of \( 50\% \), employing the median in place of the mean results in reduced sample efficiency, resulting in a drop in performance relative to averaging-based approaches in the absence of adversaries. While this fact is acknowledged in the literature~\cite{Blanchard17}, it is generally accepted as a necessary price to pay for the guarantee of robustness in the presence of adversaries. An alternative based on \( \ell_p \)-norm penalization of deviation from consensus is presented in~\cite{Li19}.

The aforementioned works~\cite{Kakade19, Yin18, Blanchard17, Li19} focus on centralized or federated learning in the presence of a fusion center. Generalizations to the decentralized setting of trimmed-mean, median and Krum based approaches have been provided in~\cite{Fang22, Bajwa20}, and of the penalty based RSA-approach in~\cite{Peng20}. We note that other works, such as~\cite{Sara17}, have considered the problem of distributed robust estimation by networked agents. Here, a collection of benign agents, all following a prescribed learning protocol, aim to learn collaboratively from contaminated data. Robustness in this context is achieved by adjusting the update~\eqref{eq:adapt}, rather than the aggregation scheme~\eqref{eq:combine}.

\section{M- and MM-based Aggregation}
Both~\eqref{eq:general_aggregation} and~\eqref{eq:median_aggregation} can be viewed as instances of the more general M-estimation problem~\cite{Marrona06, Zoubir18}:
\begin{align}\label{eq:robust_loss}
  \w_{k, i} = \argmin_{\w} \sum_{\ell \in \mathcal{N}_k} a_{\ell k} \rho^{\mathrm{agg}}\left(\boldsymbol{\phi}_{\ell, i} - \w\right)
\end{align}
The choice \( \rho^{\mathrm{agg}}(\cdot) = \|\cdot\|^2 \) yields the ordinary average, with high efficiency, but low robustness, while the choice \( \rho^{\mathrm{agg}}(\cdot) = \|\cdot\| \) yields the geometric median, with high robustness, but low efficiency. Letting \( \rho^{\mathrm{agg}}(\cdot) = \|\cdot\|_1 \) on the other hand yields the elementwise median. Different choices of \( \rho^{\mathrm{agg}}(\cdot) \) allow for the trade-off of robustness and efficiency. For simplicity, we will be focusing on loss functions \( \rho^{\mathrm{agg}}(\cdot) \), which operate elementwise on their argument, which will in turn translate into elementwise aggregation schemes. For such \( \rho^{\mathrm{agg}}(\cdot) \), we have:
\begin{align}
  \sum_{\ell \in \mathcal{N}_k} a_{\ell k} \rho^{\mathrm{agg}}\left(\boldsymbol{\phi}_{\ell, i} - \w\right) = \sum_{\ell \in \mathcal{N}_k} a_{\ell k} \sum_{m=1}^M \rho\left(\boldsymbol{\phi}_{\ell, i}(m) - \w(m)\right)
\end{align}
Popular choices for the penalty function \( \rho(\cdot) \) include monotone choices such as the Huber loss and redescending ones such as the \textcolor{black}{Tukey's} bisquare function --- for a detailed discussion on robust loss functions for location estimation we refer the reader to~\cite{Marrona06}. An alternative formulation of~\eqref{eq:robust_loss} follows after differentiating:
\begin{align}\label{eq:robust_derivatives}
  \sum_{\ell \in \mathcal{N}_k} a_{\ell k} \psi\left(\boldsymbol{\phi}_{\ell, i}(m) - \w_{k, i}(m)\right) = 0
\end{align}
where \( \psi(\cdot) = \rho'(\cdot) \) is the derivative of the loss. If we define:
\begin{align}
  b(y) \triangleq \begin{cases} \frac{\psi(y)}{y} \ &\textrm{if } y \neq 0, \\ \psi'(0) \ &\textrm{if } y = 0. \end{cases}
\end{align}
it follows that after algebraic manipulation that~\cite{Marrona06}:
\begin{align}
  \w_{k, i}(m) = \frac{\sum_{\ell \in \mathcal{N}_k} a_{\ell k} b\left(\boldsymbol{\phi}_{\ell, i}(m) - \w_{k, i}(m)\right) \boldsymbol{\phi}_{\ell, i}(m)}{\sum_{\ell \in \mathcal{N}_k} a_{\ell k} b\left(\boldsymbol{\phi}_{\ell, i}(m) - \w_{k, i}(m)\right)}
\end{align}
If we define:
\begin{align}
  \overline{\boldsymbol{a}}_{\ell k}(m) \triangleq \frac{a_{\ell k}b\left(\boldsymbol{\phi}_{\ell, i}(m) - \w_{k, i}(m)\right)}{\sum_{\ell \in \mathcal{N}_k} a_{\ell k} b\left(\boldsymbol{\phi}_{\ell, i}(m) - \w_{k, i}(m)\right)}
\end{align}
this gives rise to the representation:
\begin{align}\label{eq:robust_convex}
  \w_{k, i}(m) = \sum_{\ell \in \mathcal{N}_k} \overline{\boldsymbol{a}}_{\ell k}(m) \boldsymbol{\phi}_{\ell, i}(m)
\end{align}
Relation~\eqref{eq:robust_convex} indicates that robust aggregation via M-estimation can be interpreted as a convex combination of prior estimates \( \boldsymbol{\phi}_{\ell, i}(m) \) with weights \( \overline{\boldsymbol{a}}_{\ell k}(m) \), which are obtained by modulating \( a_{\ell k} \) with \( b\left(\boldsymbol{\phi}_{\ell, i}(m) - \w_{k, i}(m)\right) \). Here, \( b\left(\boldsymbol{\phi}_{\ell, i}(m) - \w_{k, i}(m)\right) \) measures the likelihood that the estimate obtained from neighbor \( \ell \) is an outlier. It is worth noting that while~\eqref{eq:robust_convex} indicates that \( \w_{k, i}(m) \) is a convex combination of \( \boldsymbol{\phi}_{\ell, i}(m) \), this relationship is not prescriptive, nor does it imply that it is linear. This is because \( \overline{\boldsymbol{a}}_{\ell k}(m) \) is an implicit function of the prior estimates \( \boldsymbol{\phi}_{\ell, i}(m) \) as well as the resulting estimate \( \w_{k, i}(m) \). In practice, M-estimates are pursued by fixed-point iterations, which return the weights \( \overline{\boldsymbol{a}}_{\ell k}(m) \) as a byproduct -- we refer the reader to~\cite{Marrona06} for details.

Classical M-estimators trade off robustness and statistical efficiency via the choice of the loss function \( \rho(\cdot) \). Simultaneous robustness and efficiency can be achieved as well by utilizing a nested procedure where a robust, but not efficient, estimate of location and scale is used to initialize and normalize the fixed-point recursion of a subsequent M-estimator leading to~\eqref{eq:robust_convex}. The resulting procedure is known as MM-estimation, and preserves the robustness of the initialization, while inheriting the statistical efficiency of the subsequent M-estimation~\cite{Marrona06}. In particular, MM-estimators can exhibit tolerance of close to 50\% outliers, while having efficiency close to that of the maximum likelihood estimate. We can then integrate the MM-based aggregator into our distributed learning framework to obtain the proposed algorithm\textcolor{black}{, termed REF-Diffusion for ``Robust-and -Efficient Diffusion''}:
\begin{algorithm}
  \SetAlgoLined%
  \BlankLine
  \textbf{Step 1:} At each agent \( k \), collect \( \x_{k, i} \) and update:
  \begin{align}
    \boldsymbol{\phi}_{k,i} &= \w_{k,i-1} - \mu \widehat{\nabla J}_{k}(\w_{k,i-1})\label{eq:adapt_algo}
  \end{align}
  \BlankLine
  \textbf{Step 2:} Collect \( \left\{ \boldsymbol{\phi}_{\ell,i} \right\}_{\ell \in \mathcal{N}_k} \), and compute \( \overline{\boldsymbol{a}}_{\ell k}(m) \) for \( m = 1, \ldots, M \) using a robust and efficient MM-procedure.\\
  \BlankLine
  \textbf{Step 3:} Aggregate via~\eqref{eq:robust_convex} for \( m = 1, \ldots, M \).
  \caption{\textcolor{black}{REF}-Diffusion Strategy}\label{alg:only}
\end{algorithm}

\section{Analysis}
\subsection{Modeling Conditions}
The set of agents \( \mathcal{N} \) is decomposed into two sets. The collection of benign agents is denoted by \( \mathcal{N}^b \), while the set of malicious agents is denoted by \( \mathcal{N}^m \). Benign agents in \( \mathcal{N}^b \) follow the learning and aggregation procedures in Algorithm~\ref{alg:only} faithfully, while agents in \( \mathcal{N}^m \) may deviate arbitrarily. For each agent \( k \), we similarly denote by \( \mathcal{N}_k^b \) the benign agents within the neighborhood \( \mathcal{N}_k \) of agent \( k \), and by \( \mathcal{N}_k^m \) the malicious agents within that same neighborhood.
\begin{assumption}[\textbf{Contamination Rate}]\label{as:contamination}
  For each benign agent \( k \in \mathcal{N}^b \), the majority of agents in its neighborhood are benign. Specifically:
  \begin{align}\label{eq:contamination}
    \frac{\left| \mathcal{N}_k^b \right|}{\left| \mathcal{N}_k \right|} > 1 - \epsilon
  \end{align}
  Here, \( |\cdot| \) denotes the cardinality of a set, and \( 0 \le \epsilon < \frac{1}{2} \) represents an upper bound on the fraction of malicious agents. Furthermore, the collection of benign agents \( \mathcal{N}^b \) form a connected subgraph of the full network \( \mathcal{N} \). \hfill\qed
\end{assumption}
Assumption~\eqref{as:contamination} ensures that the majority of agents within each neighborhood are benign, and that the remaining network after removing malicious agents remains connected. Such conditions are standard in the development of robust decentralized algorithms~\cite{Peng20}. Next, we introduce a condition on the MM-estimator:
\begin{assumption}[\textbf{Robust Aggregator}]\label{as:aggregator}
  The MM-estimator yielding the weights \( \overline{\boldsymbol{a}}_{\ell k}(m) \) is robust and efficient with breakdown points greater than \( \epsilon \). \hfill\qed
\end{assumption}
Finally, we impose standard conditions on the loss functions of benign agents as well as the accuracy of the gradient approximation \( \widehat{\nabla J}_{k}(\w_{k,i-1}) \):
~\cite{Sayed14proc,Sayed14, Chen15transient}:
\begin{assumption}[\textbf{Lipschitz Gradients}]\label{as:lipschitz_gradient_general_regularizer}
  For each \(k\), the gradient \(\nabla J_k(\cdot)\) is Lipschitz, namely, there exists \( \delta \geq 0 \) such that for any \(x,y \in \mathds{R}^{M}\):
  \begin{equation}\label{eq:gradbound}
    \|\nabla J_k(x) - \nabla J_k(y)\| \le \delta \|x-y\|
  \end{equation}\hfill\qed
\end{assumption}
\begin{assumption}[\textbf{Strong Convexity}]\label{as:strongcon}
  For each \(k\), the cost \(J_k(\cdot)\) is \( \nu \)-strongly convex, i.e., for every \(x,y \in \mathds{R}^{M}\):
  \begin{equation}\label{eq:jkstrong}
    {\left(x-y\right)}^{\T} \left(\nabla J_k(x) - \nabla J_k(y)\right) \ge \nu \|x-y\|^2
  \end{equation}\hfill\qed
\end{assumption}
\begin{assumption}[\textbf{Gradient Noise Process}]\label{as:noise}
  For each \( k \), the gradient noise process is defined as
  \begin{equation}
    \s_{k,i}(\w_{k,i-1}) = \widehat{\nabla J}_k(\w_{k,i-1}) - \nabla J_k(\w_{k,i-1})
  \end{equation}
  and satisfies
    \begin{align}
      \E \left[ \s_{k,i}(\w_{k,i-1}) | \boldsymbol{\mathcal{F}}_{i-1} \right] &= 0\\
      \E \left[ \|\s_{k,i}(\w_{k,i-1})\|^2 | \boldsymbol{\mathcal{F}}_{i-1} \right] &\le \beta^2 \|w^o - \w_{k,i-1}\|^2 + \sigma^2 \label{eq:noisebound}
    \end{align}
  for some non-negative constants \( \{\beta^2,\sigma^2\} \), and where \( \boldsymbol{\mathcal{F}}_{i-1} \) denotes the filtration generated by the random processes \( \{\w_{\ell,j}\} \) for all \( \ell=1,2,\ldots,K \) and \( j \le i-1 \).\hfill\qed%
\end{assumption}

\subsection{Convergence Analysis}
Assumptions 1 and 2 ensure that the number of malicious agents within each neighborhood is smaller than the breakdown point of the MM-estimator driving the aggregation procedure. This ensures that the aggregate \( \w_{k, i} \) obtained from~\eqref{eq:robust_convex} provides a meaningful estimate of the mean of \( \left\{\boldsymbol{\phi}_{\ell,i}\right\}_{\ell \in \mathcal{N}_k^b} \) over the set of \emph{benign} agents. Specifically, one expects for an efficient estimator that:
\begin{align}\label{eq:critical_approximation}
  b\left(\boldsymbol{\phi}_{\ell, i}(m) - \w_{k, i}(m)\right) \approx \begin{cases} 1, \ \textrm{if } \ell \in \mathcal{N}_k^b, \\ 0, \ \textrm{if } \ell \in \mathcal{N}_k^m. \\ \end{cases}
\end{align}
This translates to:
\begin{align}
  \overline{\boldsymbol{a}}_{\ell k}(m) \approx \overline{a}_{\ell k} \triangleq \begin{cases} \frac{a_{\ell k}}{\sum_{\ell \in \mathcal{N}_k^b} a_{\ell k}}, \ \textrm{if } \ell \in \mathcal{N}_k^b, \\ 0, \ \textrm{if } \ell \in \mathcal{N}_k^m. \\ \end{cases}
\end{align}
In other words, the effective weights \( \overline{a}_{\ell k} \) of benign agents are obtained by scaling the original weights \( a_{\ell k} \), to account for the fact that the effective weights \( \overline{a}_{\ell k} \) of malicious agents are set to zero. This ensures that effective weights continue to add up to one. Under this approximation, we can write Algorithm~\ref{alg:only} as:
\begin{align}
  \boldsymbol{\phi}_{k,i} =&\: \w_{k,i-1} - \mu \widehat{\nabla J}_{k}(\w_{k,i-1})\label{eq:approx_adapt}\\
  \w_{k,i} \approx&\: \sum_{\ell \in \mathcal{N}_{k}^b} \overline{a}_{\ell k} \boldsymbol{\phi}_{k,i-1}\label{eq:approx_combine}
\end{align}
Comparing~\eqref{eq:approx_adapt}--\eqref{eq:approx_combine} with the classical diffusion strategy~\eqref{eq:adapt}--\eqref{eq:combine}, we note two differences. First, the aggregation step~\eqref{eq:approx_combine} involves averaging only over the set of benign agents \( \mathcal{N}_k^b \) within \( \mathcal{N}_k \), and second the weights \( \overline{a}_{\ell k} \) are adjusted from \( a_{\ell k} \). The adjacency matrix \( [\overline{A}]_{\ell k} \triangleq \overline{a}_{\ell k}\) can be decomposed as:
\begin{align}
  \overline{A} = \begin{pmatrix} \overline{A}^b & 0 \\ 0 & 0 \end{pmatrix}
\end{align}
where \( \overline{A}^b \) contains the weights \( \overline{a}_{\ell k} \) of benign agents \( \ell \in \mathcal{N}^b \). Assumption 1, in light of the Perron-Frobenius theorem~\cite{Horn03}, then ensures that \( \overline{A}^b \) is a \emph{primitive} matrix with a single eigenvalue at one and corresponding eigenvector \( \overline{p}^b \), which can be normalized to satisfy:
\begin{align}
  \overline{A}^b \overline{p}^b = \overline{p}^b, \ \ \ \overline{p}^b(k) > 0 \ \forall \ k, \ \ \ \sum_{k \in \mathcal{N}^b} \overline{p}^b(k) = 1
\end{align}
We can then appeal to known results on the convergence of the non-robust diffusion strategy~\cite[Theorem 9.1]{Sayed14} to conclude:
\begin{theorem}[\textbf{Limiting Behavior}]
  Suppose Assumptions~\ref{as:contamination}--\ref{as:noise} hold, and the approximation~\eqref{eq:critical_approximation} is accurate. Then, the limiting point of Algorithm~\ref{alg:only} is determined by the data \( \x_k \) of \emph{benign} agents \( \mathcal{N}^b \) through:
  \begin{align}
    \overline{w}^o \triangleq \argmin_w \sum_{k \in \mathcal{N}^b} \overline{p}^b(k) \mathds{E} Q(w; \x_k)
  \end{align}
  We have for all \( k \in \mathcal{N}^b \):
  \begin{align}
    \limsup_{i \to \infty} \mathds{E} \|\overline{w}^o - \w_{k, i} \|^2 = O(\mu)
  \end{align}
  for sufficiently small step-size \( \mu \).
\end{theorem}

\section{Numerical Results}
We consider a collection of \( K = {32} \) agents, connected through a fully connected graph. Each agent observes data following a linear model:
\begin{align}
  \boldsymbol{d}_k = \boldsymbol{u}_k^{\mathsf{T}} w^o + \boldsymbol{v}_k
\end{align}
where the regressors \( \boldsymbol{u}_k \in \mathds{R}^{10} \) are identically normally distributed with \( \boldsymbol{u}_k \sim \mathcal{N}(0, {I_{10}}) \). The noise term \( \boldsymbol{v}_k \) is also normally distributed with \( \boldsymbol{v}_k \sim \mathcal{N}(0, \sigma_v^2) \) and \( \sigma_v^2 = 0.01 \). Each agent is equipped with the mean square error cost:
\begin{align}
  J_k(w) = \frac{1}{2}\mathds{E} \|\boldsymbol{d}_k - \boldsymbol{u}_k^{\mathsf{T}} w\|^2
\end{align}
and constructs the gradient approximation:
\begin{align}
  \widehat{\nabla J}_k(w) \triangleq \boldsymbol{u}_k \left( \boldsymbol{d}_k - \boldsymbol{u}_k^{\mathsf{T}} w \right)
\end{align}
It can be readily verified that this formulation satisfies Assumption~\ref{as:lipschitz_gradient_general_regularizer} through~\ref{as:noise}. Benign agents follow the prescribed learning and aggregation schemes. The proposed scheme of Algorithm~\ref{alg:only} is implemented through an M-estimator with Tukey's biweight loss function~\cite{Marrona06}, initialized and normalized with robust location and scale estimates through the median and median absolute deviation respectively. The implementation is taken from the repository of~\cite{Zoubir18}, available publicly on Github. Performance is compared to the baseline averaging-based approach~\cite{Sayed14} and elementwise median aggregation~\cite{Yin18}. A variable number of malicious agents deviate from the prescribed learning protocol by additively perturbing their local update via:
\begin{align}
  \boldsymbol{\phi}_{k,i} &= \w_{k,i-1} - \mu \widehat{\nabla J}_{k}(\w_{k,i-1}) + \boldsymbol{\Delta}
\end{align}
where \( \boldsymbol{\Delta} = \delta \mathds{1} \).

\begin{figure}
\begin{subfigure}[b]{.49\linewidth}
  \centering
  \includegraphics[width=\linewidth]{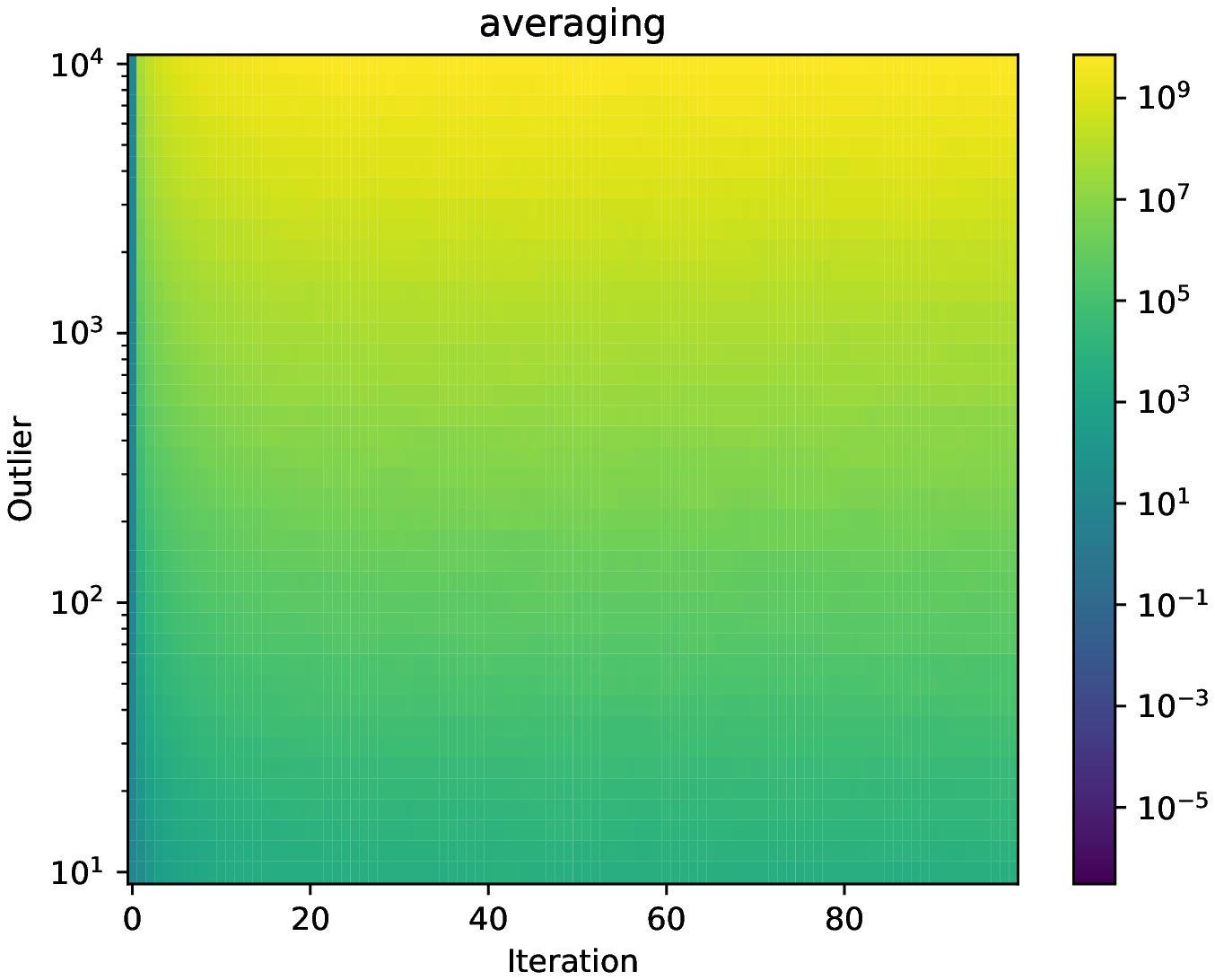}
  \includegraphics[width=\linewidth]{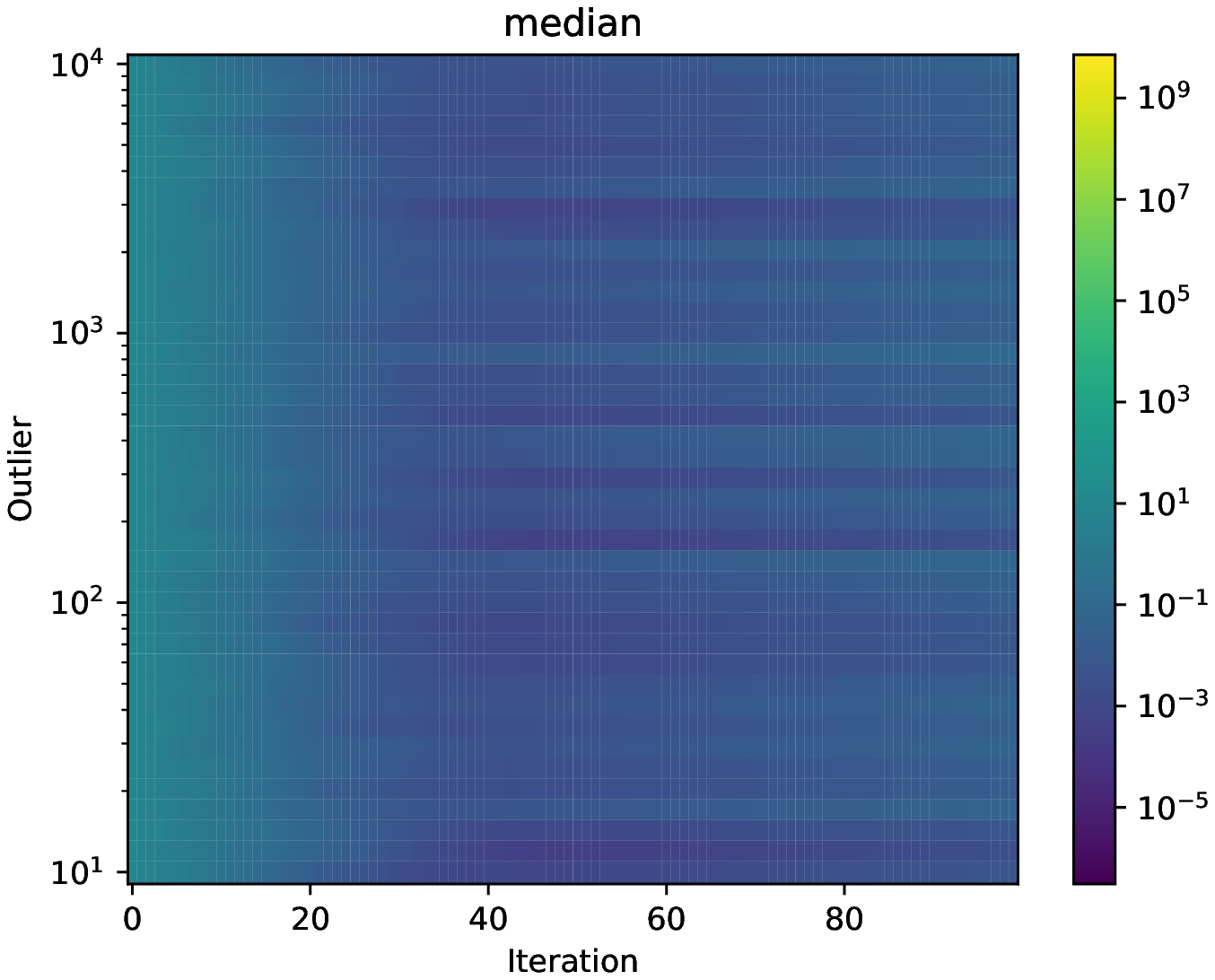}
  \includegraphics[width=\linewidth]{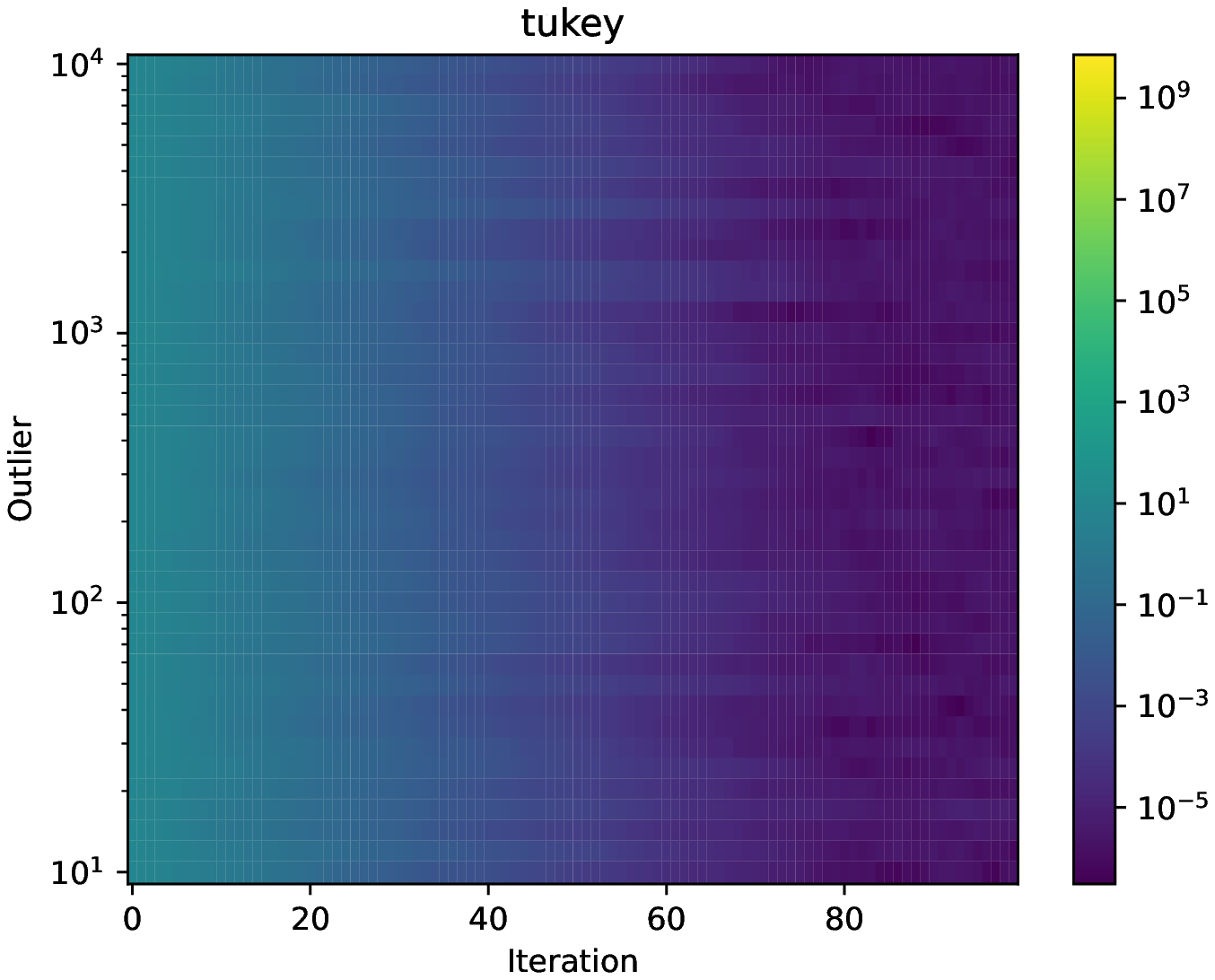}
  \end{subfigure}
  \begin{subfigure}[b]{.49\linewidth}
  \centering
  \includegraphics[width=\linewidth]{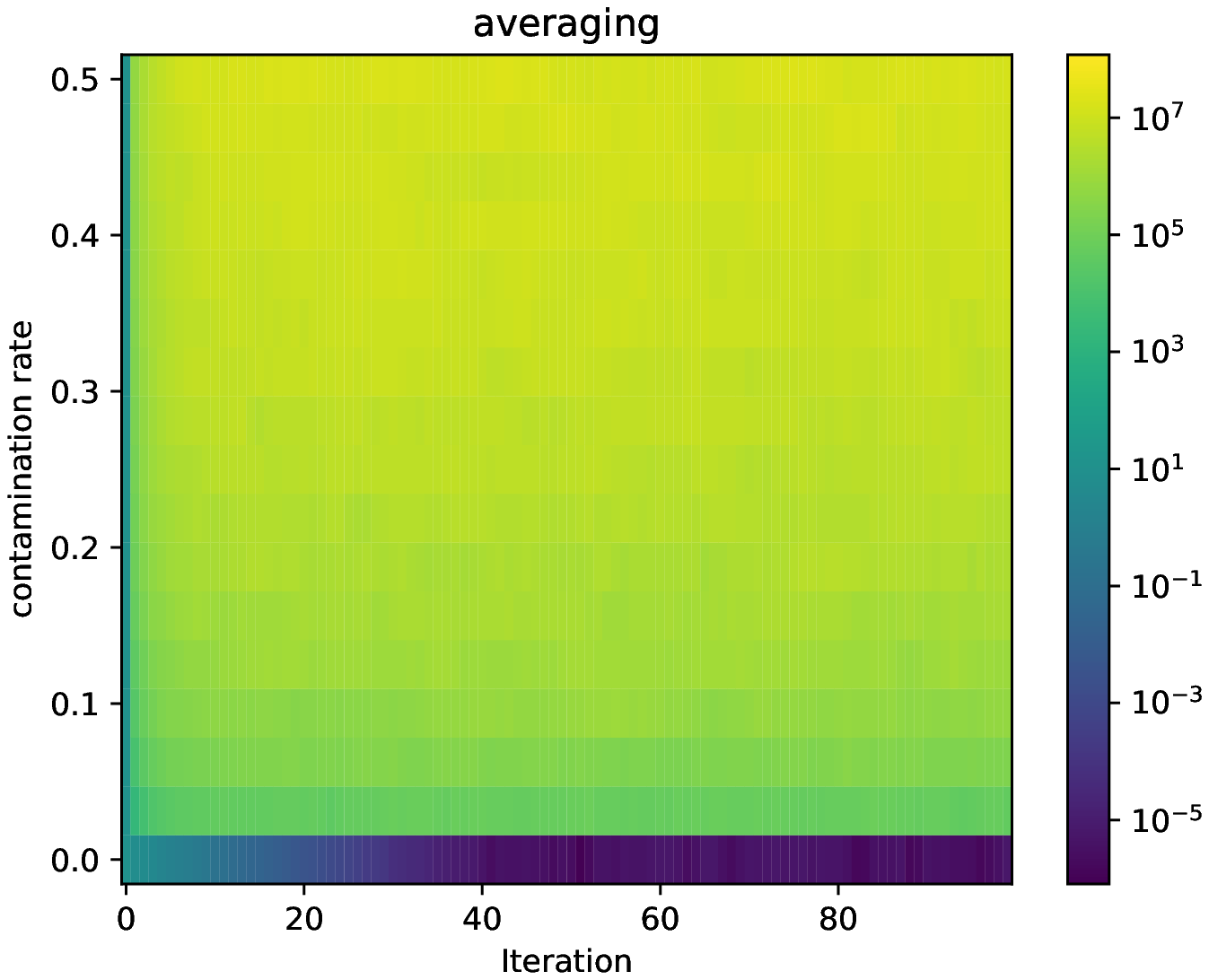}
  \includegraphics[width=\linewidth]{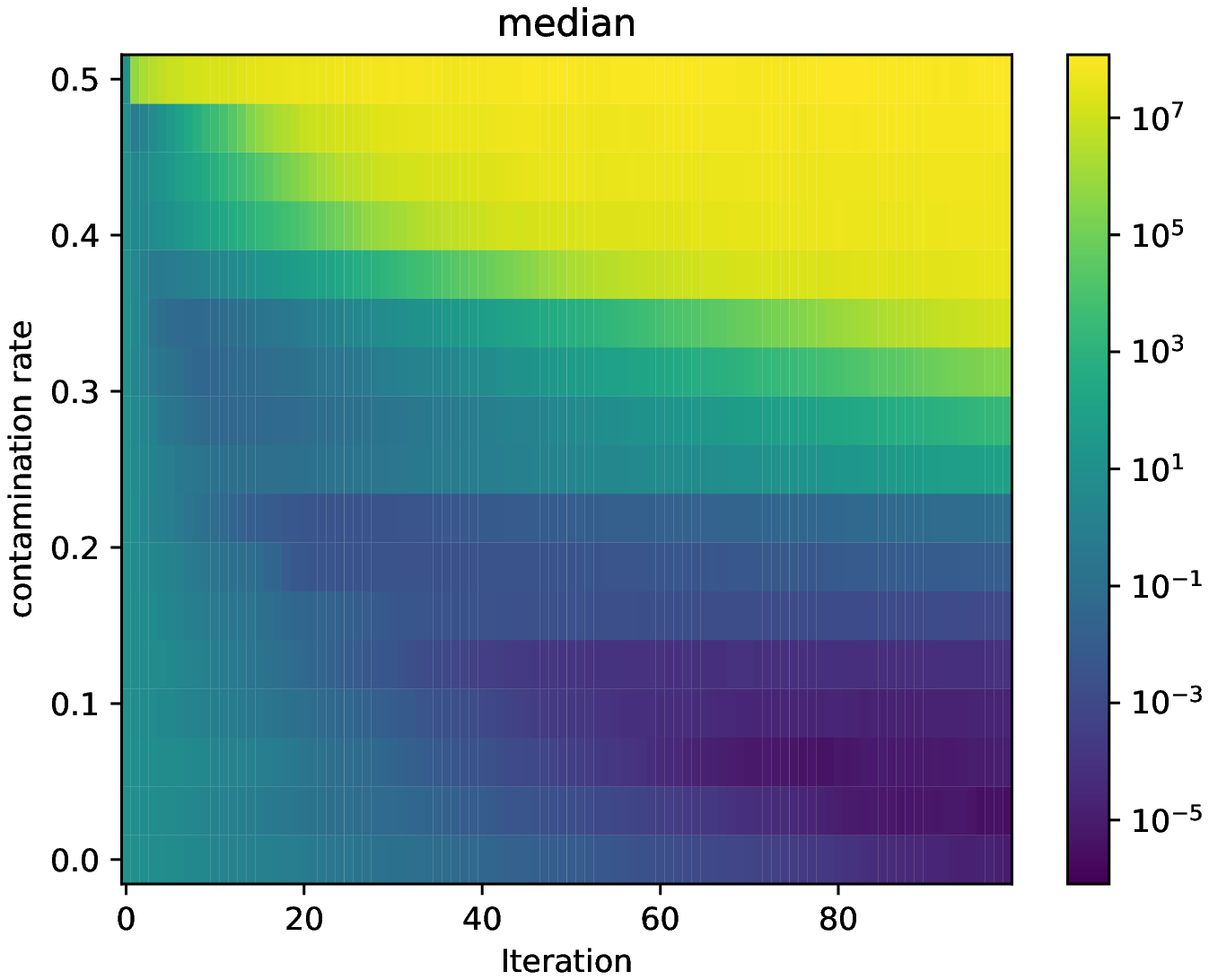}
  \includegraphics[width=\linewidth]{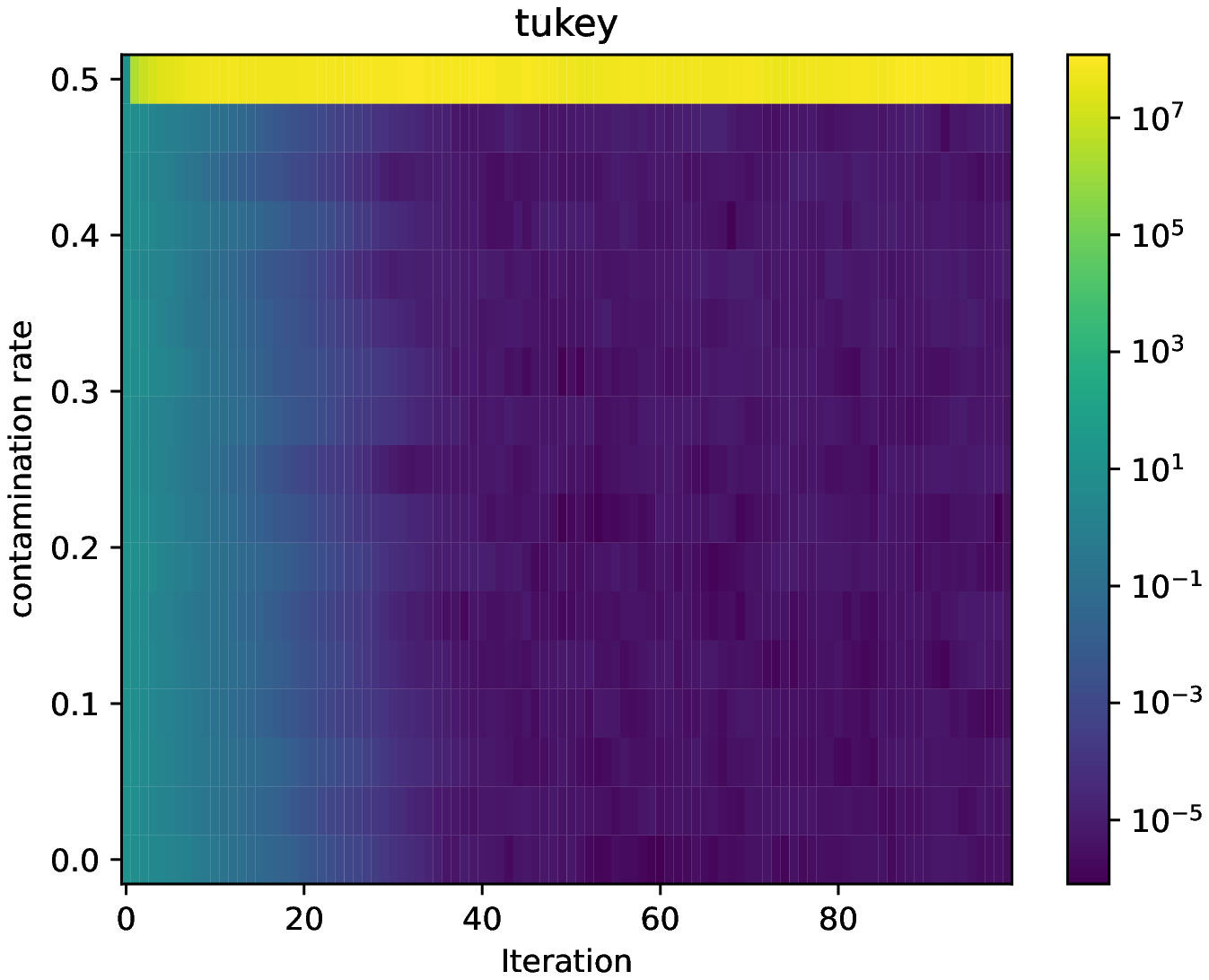}
  \end{subfigure}
  \caption{Performance over time and contamination strength \( \delta \) for a single malicious agent (left) and performance over time and contamination rate for a fixed strength (right).}\label{fig:varying}
\end{figure}

We show in in the left column of Fig.~\ref{fig:varying} the mean-square deviation from \( w^o \) for a single malicious agent, as a function of both iteration and contamination strength \( \delta \). In the right column of Fig.~\ref{fig:varying} we show mean-square deviation for a fixed contamination strength \( \delta = 1000 \) as a function of both iteration and rate of contamination.

\section{Conclusion}
We have presented \textcolor{black}{REF}-Diffusion, an algorithm for robust and efficient learning over networks. The strategy is derived by replacing traditional averaging- or median-based aggregation procedures by an MM-estimate of location, which can be designed to be simultaneously robust and efficient. The result is a strategy which performs on par with averaging-based approaches in the absence of deviating agents, while preserving robustness in the presence of perturbations. Numerical results corroborate the claims.

\bibliographystyle{IEEEbib}
{\bibliography{main}}

\end{document}